\pdfoutput=1
\documentclass[11pt]{article}

\usepackage[final]{acl}

\usepackage{times}
\usepackage{latexsym}

\usepackage[T1]{fontenc}
\usepackage[utf8]{inputenc}

\usepackage{microtype}

\usepackage{inconsolata}
\usepackage{amsmath}

\usepackage{graphicx}

\usepackage{xcolor}  % グラフィックス関連
\usepackage{pxrubrica}        % ルビ
\usepackage{url}
\usepackage{booktabs} 
\usepackage{hyperref}
\hypersetup{
	colorlinks=true, 
    citecolor=blue, 
    linkcolor=blue,
    urlcolor=blue,
	pdfborder={0 0 0},
}
\usepackage{xcolor}
\usepackage{tcolorbox}
\usepackage[verb]{bxghost}    % \verb 前後に適切な和欧文間スペース
\usepackage{listings}
\usepackage{tcolorbox}
\usepackage{xcolor}

\definecolor{commentcolor}{RGB}{0,100,0}   % Dark green for comments
\definecolor{keywordcolor}{RGB}{0,0,255}   % Blue for keywords
\definecolor{stringcolor}{RGB}{163,21,21}  % Dark red for strings
\definecolor{identifiercolor}{RGB}{0,0,0}  % Black for identifiers

\newtcolorbox{mybox}[1][]{
    title=#1,
    fonttitle=\small,
    fontupper=\small,
    left=2mm,
    right=2mm,
    top=1mm,
    bottom=0mm,
}

\lstdefinestyle{mystyle}{
    commentstyle=\color{commentcolor},
    keywordstyle=\color{keywordcolor}\bfseries,
    stringstyle=\color{stringcolor},
    identifierstyle=\color{identifiercolor},
    basicstyle=\ttfamily\lst@ifdisplaystyle\tiny\fi,
    breakatwhitespace=false,
    breaklines=true,
    captionpos=b,
    keepspaces=true,
    numbers=none,
    numbersep=5pt,
    showspaces=false,
    showstringspaces=false,
    showtabs=false,
    tabsize=2,
    xleftmargin=0pt,
}
\title{An Empirical Study of LLM-as-a-Judge: \\ How Design Choices Impact Evaluation Reliability}

\author{Yusuke Yamauchi\thanks{Equal contribution.}\thanks{Work done during an internship at NEC Corporation.} \\
  The University of Tokyo   \\
  \texttt{y\_yamauchi@is.s.u-tokyo.ac.jp} \\\And
  Taro Yano\footnotemark[1] \\
  NEC Corporation \\ \texttt{taro\_yano@nec.com} \\\And
  Masafumi Oyamada \\
  NEC Corporation \\ \texttt{oyamada@nec.com}
 \\}

\begin{document}
\maketitle
\begin{abstract}
% In recent years, large language models (LLMs) have exhibited remarkable performance across various tasks, significantly impacting both society and academia. Among the various evaluation methods for LLMs, open-ended evaluation is particularly important for measuring their capabilities and instruction-following abilities, especially in chat assistant applications. The LLM-as-a-Judge approach has gained attention as a cost-effective and efficient alternative to human evaluation. However, challenges remain regarding the stability of the evaluation. 
% In this study, we examine the phenomenon in which evaluation scores fluctuate even when the evaluator LLM is presented with the same instruction and response.

% Building on our findings, we propose the hypothesis that \emph{ambiguity in evaluation criteria induces score fluctuations}. To test this hypothesis, we intentionally introduce ambiguity into the evaluation criteria and observe the resulting fluctuations in scores. Our analysis reveals that the higher the degree of ambiguity in the evaluation criteria, the greater the score fluctuation. 
% Unlike previous research focusing on biases in evaluator LLMs, this study demonstrates that score fluctuation arises not only from the evaluator LLM itself but also from the design of the evaluation task.
As large language models (LLMs) continue to advance, reliable evaluation methods are essential—particularly for open-ended, instruction-following tasks. LLM-as-a-Judge enables automatic evaluation using LLMs as evaluators, but its reliability remains uncertain. In this work, we analyze key factors affecting its trustworthiness, focusing on alignment with human judgments and evaluation consistency. Using BIGGENBench and EvalBiasBench, we study the effects of evaluation design, decoding strategies, and Chain-of-Tought (CoT) reasoning in evaluation. Our results show that evaluation criteria are critical for reliability, non-deterministic sampling improves alignment with human preferences over deterministic evaluation, and CoT reasoning offers minimal gains when clear evaluation criteria are present.
\end{abstract}

\section{Introduction}
In recent years, large language models (LLMs) have been evolving rapidly, demonstrating high performance across various tasks~\cite{DBLP:journals/corr/abs-2303-08774,claude35sonnet, gemini1_5} and exerting significant influence. In addition to the high-performing proprietary models, there have been active efforts to develop open, small and high-performance LLMs~\cite{DBLP:journals/corr/abs-2407-21783, DBLP:journals/corr/abs-2412-15115, DBLP:journals/corr/abs-2412-19437, DBLP:journals/corr/abs-2404-14219}.  

To compare these LLMs, it is necessary to evaluate their performance on various tasks. Open-ended evaluation is particularly required to measure response capabilities and instruction-following ability as chat assistants. LLM-as-a-Judge~\cite{mt-bench} is a technique developed for open-ended evaluation, where an evaluator LLM measures the performance of benchmarked LLMs. This approach has the advantage of being lower-cost and faster than manual evaluation~\cite{DBLP:journals/corr/abs-2411-15594}.  

However, despite its growing adoption, there remain open questions about the \textit{reliability} of LLM-as-a-Judge. In particular, we investigate two essential properties to ensure its trustworthiness: 1. \textbf{Alignment with human judgments}~\citep{DBLP:journals/corr/abs-2412-05579}, and 2. \textbf{Consistency of evaluation results}~\citep{DBLP:journals/corr/abs-2412-12509, DBLP:journals/corr/abs-2408-13006}.
Without these properties, automatic evaluation using LLMs risks producing misleading conclusions about model performance.

In this work, we aim to identify key factors that affect the reliability of LLM-as-a-Judge. To this end, we conduct a series of empirical analyses using two public benchmarks—\textbf{BIGGENBench}~\citep{BiGGEN-Bench} and \textbf{EvalBiasBench}~\citep{park2024offsetbiasleveragingdebiaseddata}—which provide a diverse set of open-ended tasks. Through systematic experiments, 
we investigate the impact of 1. the presence or absence of \textbf{reference answers} and \textbf{score descriptions} in evaluation prompts, 2. the choice of \textbf{decoding strategy} (greedy vs.\ sampling) used by the evaluator model, and 3. the role of \textbf{CoT} in the evaluator's response.

Our findings reveal that:
\begin{enumerate}
  \item \textbf{Evaluation design:} Providing both reference answers and score descriptions is crucial for reliable evaluation. Omitting either significantly degrades alignment with human judgments, especially for weaker evaluator models. Furthermore, providing descriptions only for the highest and lowest scores yields the most reliable results, suggesting that the necessity of descriptions for intermediate scores should be reconsidered.
  
  \item \textbf{Decoding strategy:} Greedy decoding ensures zero score variance, but it tends to show lower correlation with human judgments compared to sampling-based decoding. Sampling introduces variability in scores, but it better captures the nuances of human preferences. Furthermore, averaging scores aligns with human judgments the most among compared three aggregation methods.
  
  \item \textbf{Use of CoT reasoning:} When well-defined score descriptions are available, including CoT reasoning in evaluator responses has little effect on alignment with human judgments. From both a cost and performance perspective, CoT-free scoring combined with score averaging provides strong alignment with human evaluations while maintaining low computational cost.
\end{enumerate}

\section{Related Work}\label{sec:related work}
% LLM-as-a-Judge は多様な正解を許容するタスクで必要
\textbf{Evaluation of LLMs.} Evaluating LLMs for generative tasks involves significant manual costs, leading to autonomous evaluation methods. Traditional approaches measure similarity between model outputs and references using lexical features (BLEU~\cite{papineni-etal-2002-bleu}, ROUGE~\cite{lin-2004-rouge}, CIDEr~\cite{vedantam2015ciderconsensusbasedimagedescription}) or semantic features (BERTScore~\cite{zhang2020bertscoreevaluatingtextgeneration}, COMET~\cite{rei2020cometneuralframeworkmt}). However, these methods struggle with tasks allowing diverse valid responses. LLM-as-a-Judge~\cite{mt-bench} addresses this by using capable models like GPT-4 as evaluators, employing Single Answer Grading (1-10 scoring) or Pairwise Evaluation (ranking multiple outputs)~\cite{doddapaneni2024findingblindspotsevaluator}.
MT-Bench~\cite{mt-bench} assesses multi-turn capabilities using Single Answer Grading with reference answers. AlpacaEval 2.0~\cite{DBLP:journals/corr/abs-2404-04475} uses Pairwise Evaluation to mitigate length bias. Arena-Hard~\cite{DBLP:journals/corr/abs-2406-11939} filters ChatbotArena prompts for quality and diversity. BIGGEN-Bench~\cite{BiGGEN-Bench} provides instance-specific criteria improving human judgment correlation. 

% LLM-as-a-Judge では人間とアラインメントが重要
\textbf{Alignment with human judgments in LLM-as-a-Judge.} Various approaches improve alignment with human judgments, including CoT reasoning, self-generated criteria, and multiple evaluations~\cite{mt-bench, zeng2024evaluating}. Other methods optimize prompts using human annotation correlation~\cite{liu2023calibratingllmbasedevaluator, liu-etal-2024-hd} or employ ensemble voting~\cite{liu2023goalorientedpromptattacksafety}. \citet{Gu2024ASO} proposed metacognitive re-evaluation for consistency.
Our study utilizes simple methodologies from a neutral standpoint to analyze the impact of evaluation design, decoding strategies, and CoT reasoning on alignment with human judgments.

% LLM-as-a-Judge では評価の一貫性が重要
\textbf{Consistency of evaluation results in LLM-as-a-Judge.} Existing studies have identified various biases where semantically unchanged modifications affect evaluation results. \citet{DBLP:conf/emnlp/ChenCLJW24} examined gender, authority, and aesthetic biases. \citet{DBLP:journals/corr/abs-2410-02736} identified 12 major latent biases including positional and self-enhancement bias. \citet{DBLP:conf/emnlp/ParkJRKC24} highlighted challenges with response length variations and content continuity. To the best of our knowledge, this is the first research to extensively investigate how much evaluation results can fluctuate depending on the design of the evaluation tasks and the evaluation strategies.

\section{Experiments}\label{sec:reult_stability}
In this section, we examine what factors affect the alignment with human judgments and consistency of evaluation results. Research Questions (RQs) we aim to investigate are as follows:  
\begin{enumerate}
    \item Which components of evaluation design facilitate improved alignment with human judgments and enhance the consistency of evaluation results?
    \item What are the advantages and disadvantages of deterministic versus non-deterministic decoding strategies?
    \item Does CoT improve alignment with human judgments and the consistency of evaluation results?
\end{enumerate}

\subsection{Experimental Method}
\begin{figure*}[t]
  \centering
\vspace{-10pt}
\begin{mybox}[Template for Evaluation Prompt]
\vspace{-9pt}
\lstset{basicstyle=\scriptsize\ttfamily}
\begin{lstlisting}
###Task Description: 
An instruction (which may include an Input), a response to evaluate, a reference answer scoring 5, and a score rubric representing evaluation criteria are provided.  
1. Write detailed feedback assessing the response strictly based on the score rubric.  
2. After the feedback, provide an integer score from 1 to 5, referring to the rubric.  
3. The output format should be: "(write feedback for criteria) [RESULT] (an integer between 1 and 5)"  
4. Do not include any additional introductions, conclusions, or explanations.  
###The instruction to evaluate:  
{instruction}  
###Response to evaluate:  
{response}  
###Reference Answer (Score 5):  
{reference answer}  
###Score Rubrics:  
[{evaluation axes}]
Score 1: {score1_description}
Score 2: {score2_description}
Score 3: {score3_description}
Score 4: {score4_description}
Score 5: {score5_description}
###Feedback: 
\end{lstlisting}
\vspace{-6pt}
\end{mybox}
\vspace{-9pt}
    \caption{Prompt template used in our experiments to evaluate responses based on provided reference answers and evaluation criteria. The evaluation criteria consist of evaluation axes, which define general evaluation principles, and score descriptions, which provide rubrics for each of the five scores (1 through 5).}
  \label{fig:eval template}
\end{figure*}

We describe the experimental methods to investigate the RQs.
% \begin{enumerate}
%     \item \textbf{Experiment1:} To examine RQ1, we use existing benchmarks and responses on HuggingFace. Then, we select the evaluator LLM, and evaluate for $n$ times with temperature $T$. To measure the fluctuation of scores, we use Krippendorff’s alpha coefficient, which is used to measure the agreement of annotations.
%     \item \textbf{Experiment2:} To investigate RQ2, we observe the fluctuation of scores when removing the reference answer and the evaluation criteria from the evaluation prompt, identifying which elements contribute to the score fluctuation.
%     \item \textbf{Experiment3:} We explore RQ3 by conducting ablation studies that remove parts of evaluation criteria: axes and score descriptions to investigate which parts affect the fluctuation most. 
%     \item \textbf{Experiment4:} We compare the score fluctuation of different two scoring methods: scoring with feedback and scoring without feedback to confirm RQ4 by modifying the prompt template for the evaluation. 
% \end{enumerate}

% To separate the influence of the reference answer, we conduct experiments based on a w/o ref setting and perform ablation on rule elements to measure fluctuation.

\textbf{Alignment with human judgments.} To measure the degree of alignment with human judgments, we compute the correlation coefficient between the scores provided by humans and those generated by an evaluator LLM.

\textbf{Consistency of evaluation results.} We use Krippendorff's alpha coefficient to evaluate consistency of evaluation results, denoted as $\alpha$.
% , calculated as:  
% \begin{eqnarray}
%     \alpha = 1 - \frac{D_o}{D_e}.
% \end{eqnarray}  
% Here, $D_o$ represents the observed disagreement between scores, while $D_e$ denotes the expected disagreement if the evaluations were random. For a detailed definition,   
The $\alpha$ value, which indicates the consistency and reliability of evaluations, is 1 for perfect agreement, 0 for random annotations, and negative for systematic disagreement (see Appendix~\ref{appendix: alpha} for details).

\textbf{Datasets.} We adopt BIGGEN-Bench~\citep{BiGGEN-Bench}, which includes nine tasks such as instruction following, tool use, and reasoning, each with detailed, hand-crafted evaluation criteria.
The evaluation template used in our experiments is shown in Figure~\ref{fig:eval template}. 
We also use EvalBiasBench~\citep{park2024offsetbiasleveragingdebiaseddata}, an instruction-following benchmark with both correct and biased answers. We generated evaluation criteria using GPT-4o-2024-08-06 to encourage lower scores for biased responses and improve consistency.
% based on our hypothesis that when the criteria explicitly state that biased responses should be penalized, the consistency of evaluation results for such responses increases. 

% \subsection{Experimental Settings}  
\textbf{Models.} We use GPT-4o-2024-08-06 as the evaluator LLM. Furthermore, considering recent studies on self-improvement~\cite{DBLP:conf/icml/YuanPCLSXW24, DBLP:conf/nips/MadaanTGHGW0DPY23} that use local LLMs as evaluators~\cite{DBLP:journals/corr/abs-2412-02674, DBLP:journals/corr/abs-2406-01297}, we also use LLaMA-3.1-70B-Instruct~\footnote{https://huggingface.co/meta-llama/Llama-3.1-70B-Instruct}~\citep{DBLP:journals/corr/abs-2407-21783} as the evaluator LLM.

% \begin{figure*}[t]
% \centering
% \scalebox{0.85}{
% \begin{tcolorbox}[colback=white, colframe=black, width=\textwidth, sharp corners]
% \textbf{Task Description:} 

% An instruction (which may include an Input), a response to evaluate, a reference answer scoring 5, and a score rubric representing evaluation criteria are provided.  

% 1. Write detailed feedback assessing the response strictly based on the score rubric.  
% 2. After the feedback, provide an integer score from 1 to 5, referring to the rubric.  
% 3. The output format should be: "(write feedback for criteria) [RESULT] (an integer between 1 and 5)"  
% 4. Do not include any additional introductions, conclusions, or explanations.  

% \textbf{The instruction to evaluate:}  

% \{instruction\}  

% \textbf{Response to evaluate:}  

% \{response\}  

% \textbf{Reference Answer (Score 5):}  

% \{reference answer\}  

% \textbf{Score Rubrics:}  

% \{rubric\}  

% \textbf{Feedback:}  
% \end{tcolorbox}
% }
% \caption{Prompt template used in the LLM-as-a-Judge experiment, designed to evaluate responses based on provided reference answers and score rubrics.}\label{fig:eval template}
% \end{figure*}

\subsection{Results}\label{subsec:results}  

\begin{table}[h]
\begin{center}
\caption{Experimental results for RQ1 report Krippendorff's alpha coefficients across five sampled scores, with values in parentheses indicating the correlation with human evaluation. Removing evaluation criteria (\textit{w/o crt}) or reference answers (\textit{w/o ref}) reduces human correlation. Eliminating both (\textit{w/o ref\&crt}) increases score fluctuation and significantly lowers human correlation.}
\begin{tabular}{lll} 
    \toprule
    & \multicolumn{2}{c}{BIGGEN-Bench}\\
    Method&GPT-4o&LLaMA3.1\\
    \hline
    Default &0.908 (0.666) &0.806 (0.641) \\
    w/o crt &0.909 (0.591) &0.807 (0.555) \\
    w/o ref &0.921 (0.638) &0.824 (0.581) \\
    w/o ref\&crt &0.896 (0.487) &0.758 (0.346) \\
    \midrule
    & \multicolumn{2}{c}{EvalBiasBench} \\
    Method&GPT-4o&LLaMA3.1\\
    \hline
     Default  &0.865 & 0.768\\
      w/o crt  &0.839 & 0.725\\
      w/o ref  &0.869 &0.787\\
      w/o ref\&crt  &0.811 &0.753 \\
     \bottomrule
\end{tabular}
\label{tab:main result}
\end{center}
\end{table}

% \textbf{RQ1: How much do evaluation results fluctuate when evaluating using non-zero temperature decoding?}
% The results of Experiments 1 and 2 are shown in Table~\ref{tab:main result}. 
% % The values represent Krippendorff's alpha coefficient calculated over five evaluations. 
% As shown in the results of Default method, GPT-4o’s $\alpha$ is 0.9089 for BIGGEN-Bench and 0.8658 for EvalBiasBench, and LLaMA shows 0.8061 and 0.768, respectively, indicating that evaluation results fluctuate even when the instruction, response, and evaluator LLM are identical. 
% The lower $\alpha$ for LLaMA-3.1-70B-Instruct suggests larger inconsistency as the evaluator LLM compared to GPT-4o.

% 評価基準、もしくは reference answer を取り除くと人間ジャッジメントとの相関が小さくなる。GPT-4o では0.666 --> 0.591と0.666 --> 0.638, LLaMA3.1では0.641 --> 0.555 と 0.641 -->0.581であり、評価者 LLM の種類に限らず評価基準は reference answer よりも影響が大きく、各要素を削除したときの相関係数の毀損は GPT-4o よりも LLaMA3.1-70B-Instruct の方が影響が大きい。
% 評価基準と reference answer 両方を取り除くと、人間ジャッジメントの相関は大きく低下し最小となる。
% 評価結果の一貫性については、BIGGEN-Bench では評価基準、もしくは reference answer を取り除いても一貫性は保たれるが、EvalBiasBench では評価基準の削除によって評価結果の一貫性が毀損されやすいことが分かる。これは、EvalBiasBench では、バイアスドな回答に対してスコアを下げるか下げないかの評価基準が与えられない場合は、ペナルティの有無がランダムに変わってしまっている可能性がある。よって、バイアスドな回答に対するスコアの基準を明確に定めることは、評価の一貫性を高めるために重要であることが分かる。
\textbf{RQ1. Which components of evaluation design facilitate improved alignment with human judgments and enhance the consistency of evaluation results?} As shown in Table~\ref{tab:main result}, removing either the evaluation criteria or the reference answer leads to a decrease in correlation with human judgments. For GPT-4o, the correlation drops from 0.666 to 0.591 and 0.638, respectively, while for LLaMA3.1-70B-Instruct, it drops from 0.641 to 0.555 and 0.581. This indicates that, regardless of the evaluator LLM used, the evaluation criteria have a greater impact than the reference answer. Furthermore, the degradation in correlation is more pronounced for LLaMA3.1 than for GPT-4o. When both the evaluation criteria and the reference answer are removed, the correlation with human judgment declines significantly and reaches its minimum.

Regarding evaluation consistency, in the BIGGEN-Bench dataset, removing the evaluation criteria or the reference answer does not substantially affect consistency. However, in EvalBiasBench, removing the evaluation criteria leads to a noticeable drop in consistency. This suggests that, in EvalBiasBench, the absence of explicit criteria for penalizing biased responses may result in inconsistent scoring—potentially depending on random factors. Therefore, clearly defining scoring criteria for biased responses is crucial to ensure consistent evaluation.
% \begin{figure}[t]
%     \centering
%     \includegraphics[width=\columnwidth]{figures/rq3_0a.png}
%     \caption{Results for RQ1, which examines score fluctuation when removing parts of evaluation criteria, where \textit{Desc} indicates the score descriptions. When given the score description for score 5 (\textit{Score 5}) or those for scores 1, 3, and 5 (\textit{Score 1, 3, 5}), the scores fluctuate more than without score descriptions (\textit{w/o Desc}). On the other hand, when provided the score descriptions for scores 1 and 5 (\textit{Score 1, 5}), the results show the highest score consistency. These results are consistent whether axes are provided to the evaluation criteria or not.}
%     \label{fig:rq3_0}
% \end{figure}
% \begin{figure}[t]
%     \centering
%     \includegraphics[width=\columnwidth]{figures/rq3_0b.png}
%     \caption{Additional results for RQ1, which shows correlations with human evaluations when removing parts of evaluation criteria. Similar to the results in Figure~\ref{fig:rq3_0}, \textit{Score 1, 5} shows the highest correlations with human evaluations (\textit{Takeaway 7}), while \textit{Score 5} and \textit{Score 1, 3, 5} shows better correlations compared to \textit{w/o Desc}. 
%     Again, these results are consistent whether axes are provided to the evaluation criteria or not.}
%     \label{fig:rq3_0b}
% \end{figure}
\begin{figure}[t]
    \centering
    \includegraphics[width=\columnwidth]{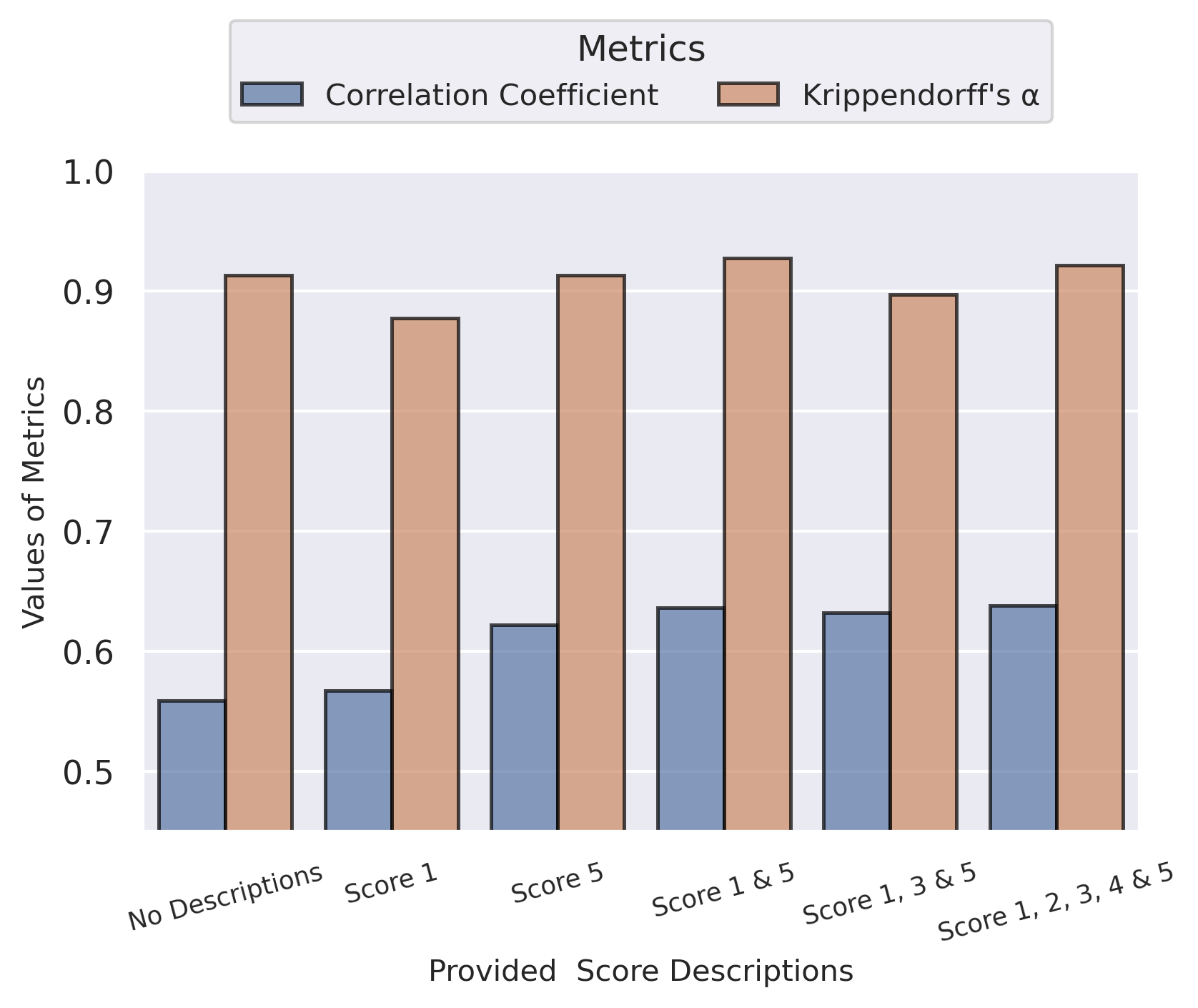}
    \caption{Additional experimental results for RQ1, showing the correlation coefficient and Krippendorff's $\alpha$ when parts of the score descriptions are removed from the evaluation criteria. When only the descriptions for scores 1 and 5 are provided (\textit{Score 1 \& 5}), the results exhibit the highest correlation with human evaluations while maintaining high evaluation consistency. This suggests that the role of score descriptions for intermediate scores (2, 3, and 4) should be reconsidered.}
    \label{fig:rq1_arxiv}
\end{figure}
% \begin{figure}[t]
%     \centering
%     \includegraphics[width=\columnwidth]{figures/rq3_1.png}
%     \caption{Results of additional experiments to investigate the question: Is it important to provide any two score descriptions or those for good and bad responses? The result shows that when the difference becomes larger, the scores become more consistent. This result supports the latter hypothesis: providing descriptions for good and bad responses is important.}
%     \label{fig:rq3_1}
% \end{figure}

Figure~\ref{fig:rq1_arxiv} illustrates the additional experimental results, which examines correleations and score fluctuation when removing a part of score descriptions from evaluation criteria. 
% スコア1と5の説明文を与えたときとすべてのスコア (1,2,3,4,5)についての説明文を与えた時で人間ジャッジとの相関もスコアの一貫性も大きく変わらない。これは intermediate のスコア (2,3,4) に対する説明文が人間ジャッジとの相関に大きな影響を与えないことを示しており、その役割について再度考える必要がある。
% すべての設定でスコアの一貫性がおおむね高い値を保っていることは意外であり、スコアの説明文が細かく与えられていなくても、一般的な評価基準である axes が与えられていれば評価は一貫する傾向にあることを示している。
The figure shows there is little difference in both correlation with human judgments and score consistency between the setting where only the descriptions for scores 1 and 5 are provided and the setting where descriptions for all scores (1, 2, 3, 4, and 5) are given. These results suggest that the descriptions for intermediate scores (2, 3, and 4) have limited impact on alignment with human judgments, and their role should be reconsidered.
It is also surprising that evaluation consistency remains generally high across all settings, indicating that even without detailed score descriptions, evaluations tend to remain consistent as long as general evaluation axes are provided.

% First, we can see that removing axes always degrades score consistency.
% The results also show that Score 1, 5, where score descriptions for scores 1 and 5 are provided, achieve the highest consistency. 
% Providing only score descriptions for score 1 leads to the highest score fluctuation, even more than providing no descriptions.
% This result raises the question: Is it important to provide any two score descriptions for consistency, or specifically to provide descriptions for good and bad responses? 
% To investigate this, we tried all ten combinations of two score descriptions, and Figure~\ref{fig:rq3_1} presents the results. Each point represents a combination, with the x-axis showing the score difference of provided descriptions and the y-axis representing Krippendorff's \(\alpha\). 
% The results indicate that providing only high, middle, or low criteria increases fluctuation while providing both high and low criteria reduces it. 
% Thus, larger score differences lead to lower fluctuation, emphasizing the importance of including the evaluation criteria for both the good and bad response criteria for consistency.

\textbf{RQ2. What are the advantages and disadvantages of deterministic versus non-deterministic decoding strategies?}

\begin{table}[h]
\begin{center}
\caption{Experimental results for RQ2. Non-deterministic scoring methods (Majority, Median, Mean) show larger correlations with human judges compared to deterministic decoding (Greedy). Among the non-deterministic methods, score averaging (Mean) shows the largest correlations with human judges consistently across different evaluator LLMs, reasoning types, and evaluation design.}
\scalebox{0.8}[0.8]{
  \begin{tabular}{lllll}
    \toprule
    Method&Default &w/o crt &w/o ref &w/o ref\&crt\\
    \hline
    \textbf{GPT-4o}\\
     Greedy&0.635 &0.571 &0.614 &0.466 \\
     Majority&0.647 &0.583 &0.627 &0.480\\
     Median&0.648 &0.581 &0.621 &0.481 \\
     Mean&\textbf{0.666} &\textbf{0.591} &\textbf{0.638} &\textbf{0.487}\\
     \midrule
    \midrule
    \multicolumn{5}{l}{\textbf{GPT-4o w/o CoT}}\\
     Greedy&0.636 &0.507 &0.612 &0.378 \\
     Majority&0.643 &0.545 &0.627 &0.406\\
     Median&0.651 &0.546 &0.629 &0.399 \\
     Mean&\textbf{0.664} &\textbf{0.570} &\textbf{0.641} &\textbf{0.422}\\
     \midrule
    \midrule
    \textbf{LLaMA3.1}\\
     Greedy&0.593 &0.524 &0.551 &0.273 \\
     Majority&0.625 &0.519 &0.555 &0.297 \\
     Median&0.624 &0.520 &0.558 &0.297 \\
     Mean&\textbf{0.641} &\textbf{0.555} &\textbf{0.581} &\textbf{0.346}\\    
    \bottomrule
  \end{tabular}
}
  \label{tab:vs greedy}
%\vspace{-6mm}
% \end{small}
\end{center}
\end{table}

We compare the correlation of scores with human judges between non-deterministic decoding and deterministic decoding on BIGGEN-Bench. For non-deterministic decoding, we sample five scores and aggregate them using majority voting (Majority), taking the median (Median), and averaging scores (Average). For deterministic decoding, we adopt greedy decoding (Greedy).

% greedy < sampling
Table~\ref{tab:vs greedy} shows the results. Non-deterministic scoring methods show larger correlations with human judges compared to deterministic decoding consistently.
This finding is consistent with the fact that, in general inference tasks, multiple sampling and aggregation of results outperforms greedy decoding~\cite{self-consistency}.
% mean が sampling でベスト
% 更に興味深いことに、non-deterministic decoding の中ではaveraging scores が評価者LLMや評価のデザイン、CoT のうむに関わらず最も高い人間との相関を見せている。
More interestingly, among non-deterministic decoding methods, averaging scores shows the highest correlation with humans regardless of the evaluator LLM, evaluation design, or presence of CoT. This can be attributed to the fact that averaging allows for expressing fine-grained nuances, such as 4.5 when an evaluator LLM is torn between scores of 4 and 5, whereas median or majority voting methods round the score to either 4 or 5, thus failing to fully leverage the LLM's capabilities as an evaluator.

\textbf{RQ3. Does CoT improve alignment with human judgments and the consistency of evaluation results?}
% \begin{table}[h]
% \begin{center}
% \caption{Results of Experiment 4 comparing score fluctuation when outputting feedback before the score (w/ CoT) versus directly outputting the score (Direct). For the Default setting, fluctuation remains similar between w/ CoT and Direct, while for w/o ref\&crt, providing feedback reduces score fluctuation.}
% \scalebox{0.85}[0.85]{
%   \begin{tabular}{lllll}
%     \toprule
%     & \multicolumn{2}{c}{BIGGEN-Bench} & \multicolumn{2}{c}{EvalBiasBench}\\
%     Method&Direct &w/ CoT &Direct &w/ CoT\\
%     \hline
%     Default &0.912 &0.908 &0.855 &0.865 \\
%     \hspace{1.5mm}w/o crt &0.818 &0.909 &0.856 &0.839 \\
%     \hspace{1.5mm}w/o ref &0.910 &0.921 &0.647 &0.869 \\
%     \hspace{1.5mm}w/o ref\&crt &0.833 &0.896 &0.650 &0.811 \\
%     \bottomrule
%   \end{tabular}
% }
%   \label{tab:feedback exp}
% \end{center}
% \end{table}
\begin{table}[h]
\begin{center}
\caption{Experimental results for RQ3. When given evaluation criteria and a reference answer (Default), scoring with CoT reasoning (w/ CoT) achieves comparable alignment with human judgments and evaluation consistency to Direct scoring (Direct).}
% \scalebox{0.85}[0.85]{
  \begin{tabular}{lll}
    \toprule
    \multicolumn{3}{c}{BIGGEN-Bench}\\
    Method&Direct &w/ CoT\\
    \hline
    Default &0.912 (0.664) &0.908 (0.666)\\
    \hspace{1.5mm}w/o crt &0.818 (0.570) &0.909 (0.591)\\
    \hspace{1.5mm}w/o ref &0.910 (0.641) &0.921 (0.638)\\
    \hspace{1.5mm}w/o ref\&crt &0.833 (0.422) &0.896 (0.487)\\
    \midrule
    \multicolumn{3}{c}{EvalBiasBench} \\
     Method&Direct &w/ CoT\\
    \hline
    Default &0.855 &0.865 \\
    \hspace{1.5mm}w/o crt &0.856 &0.839 \\
    \hspace{1.5mm}w/o ref &0.647 &0.869 \\
    \hspace{1.5mm}w/o ref\&crt &0.650 &0.811 \\  
    \bottomrule
  \end{tabular}
% }
  \label{tab:feedback exp}
\end{center}
\end{table}
% LLM-as-a-Judge における CoT の影響を調べるために、GPT-4o を使って CoT の後にスコアを出力した場合 (w/ CoT) と CoT 無しで直接スコアのみを出力した場合 (Direct) で人間ジャッジとの相関とスコアの一貫性を調べた。
To investigate the impact of CoT in LLM-as-a-Judge, we used GPT-4o to examine the correlation with human judges and the consistency of scores in two settings: one where a score was output after a Chain-of-Thought (w/ CoT), and one where only the score was output directly without any reasoning (Direct).
Table~\ref{tab:feedback exp} shows the results. In the Default setting, where evaluation criteria and reference answers are provided, both methods show similar correlation and consistency. Thus, when well-defined score descriptions are available, including explicit CoT in evaluator responses has little effect. 
From both a cost and performance perspective, direct scoring combined with score averaging provides strong alignment with human evaluations while maintaining low computational cost.

\section{Conclusion}
In this work, we conducted a comprehensive empirical analysis to identify key factors affecting the reliability of LLM-as-a-Judge. Through systematic experiments on BIGGENBench and EvalBiasBench, we found that: (1) comprehensive evaluation design with both reference answers and score descriptions is essential for human alignment; (2) sampling-based scoring with mean aggregation outperforms scoring with greedy decoding; and (3) CoT reasoning provides diminishing returns when detailed evaluation criteria are present. These findings help establish best practices for reliable automatic evaluation and provide a principled framework for LLM-as-a-Judge deployment.

\section{Limitations}
While our study provides valuable insights, we acknowledge several limitations. First, we used GPT-4o and LLaMA-3.1-70B-Instruct as evaluator LLMs, representing a closed model and an open model, respectively, and obtained consistent results. However, it remains unclear whether consistent results can be obtained when using other LLMs as evaluators.  

Additionally, the benchmarks used in this study employed different evaluation criteria: BIGGEN-Bench used human-crafted criteria, while EvalBiasBench relied on criteria generated by an LLM. Conducting experiments with evaluation criteria created using more diverse methods is an important direction. Furthermore, since all the data used in our experiments were in English, verification in other languages would also be an important work.

\bibliography{custom}

\appendix

\section{Krippendorff’s alpha coefficient}
\label{appendix: alpha}
In LLM-as-a-Judge, commonly used metrics for measuring inter-rater agreement include Pearson’s correlation coefficient, Spearman’s rank correlation coefficient, and Cohen’s Kappa \cite{bai2024benchmarking, liu2024aligning, thakur2025judgingjudgesevaluatingalignment}. However, these metrics have been criticized for reliability issues and their inability to handle various measurement scales or missing data \cite{yan2024llm-evaluator, Artstein2008SurveyAI}. Krippendorff's alpha is a general statistical measure that encompasses multiple agreement metrics and can be applied to various tasks~\cite{Krippendorff2011ComputingKA}. Krippendorff's alpha coefficient is defined as follows:

\begin{eqnarray}
    \alpha = 1 - \frac{D_o}{D_e}
\end{eqnarray}

Here, $D_o$ represents the total disagreement between pairs of ratings observed in the dataset. It is calculated using the coincidence matrices $o_{ck}$, the total sample size $N$, the total number of rating pairs $n$ for the $j$-th unit, and the difference function $\delta(c, k)$, which quantifies the discrepancy between a given rating pair $(c, k)$. The formula is expressed as:

\begin{eqnarray}
    D_o = \frac{1}{n}\sum_c\sum_k o_{ck} \delta_{ck}
\end{eqnarray}

In this study, we assume that $c, k \in \{1,2,3,4,5\}$. The coincidence matrix $o_{ck}$ is defined as:

\footnotesize
\begin{eqnarray}
    o_{ck} = \sum_j \frac{\text{Number of c-k pairs in unit} \,u}{\text{Total number of judges in unit } j - 1}
\end{eqnarray}
\normalsize

The difference function $\delta(c, k)$, assuming an interval scale for the scores, is given by:

\begin{eqnarray}
    \delta_{ck} = (c-k)^2
\end{eqnarray}

$D_e$ represents the expected disagreement under a random distribution of ratings, computed as:

\begin{eqnarray}
    D_e = \frac{1}{n(n-1)}\sum_c n_c \sum_k n_k \delta_{ck}
\end{eqnarray}

Here, $n_c$ and $n_k$ denote the respective frequencies of the ratings. The alpha coefficient approaches $1$ as different raters assign similar scores to the same unit.

\begin{figure*}[htbp]
  \centering
\vspace{-10pt}
\begin{mybox}[MT-Bench prompt template]
\vspace{-9pt}
\raggedright  % 左揃えを適用
\lstset{basicstyle=\scriptsize\ttfamily}
\begin{lstlisting}
###Task Description:
Please act as an impartial judge and evaluate the quality of the response provided by an AI assistant to the user question displayed below.
You will be given a reference answer and the assistant's answer.
Your evaluation should consider factors such as the helpfulness, relevance, accuracy, depth, creativity, and level of detail of the response. 
Begin your evaluation by providing a short explanation.
Be as objective as possible. After providing your explanation, please rate the response on a scale of 1 to 5 by strictly following this format: [RESULT] (an integer number between 1 and 5)

###The instruction to evaluate:
{instruction}

###Reference Answer:
{reference_answer}

###Assistant's Answer to evaluate:
{response}

### Feedback: 
\end{lstlisting}
\vspace{-6pt}
\end{mybox}
\vspace{-9pt}
    \caption{Prompt template used in the w/o crt setting.}
  \label{fig:mt-bench template}
\end{figure*}

\section{Experimental Details}
\subsection{Hyperprameters}
\label{appendix: experimental details}
Hyperparameters used to judge the responses are listed in Table~\ref{tab:hyperparams}. We retry the judging process while changing the seed value until a valid score is output for all samples.

\begin{table}[ht]
\begin{center}
\caption{Hyperparameters used during inference time}
\label{tab:hyperparams}
\begin{tabular}{c|c} \hline
    hyperparameter & value \\\hline
    Seed & \{0, 10, 20, 30, 40\} \\
    Max Seq Length &  8192 \\
    Temperature & 1.0 \\
    Repetition penalty & 1.03 \\\hline
\end{tabular}
\end{center}

\subsection{Response Model}
We used the responses generated by Mixtral-8x7B-Instruct-v0.1 \cite{jiang2024mixtralexperts}, Qwen1.5-7B \cite{qwen1.5}, and GPT-3.5-turbo as inputs for the Judge. 
\end{table}

\subsection{Details of Prompting Strategies}
\textbf{w/o crt prompt template.} When we remove elements from the evaluation criteria, we omit the corresponding items from the prompt template. In the w/o crt setting, we use the MT-Bench prompt \cite{mt-bench} as the base prompt.

\noindent\textbf{Direct prompt template.} When conducting the judging process in the Direct setting, only the score should be output without generating rationale. Therefore, we directly add \textbf{"\#\#\# Feedback: [Result]"} at the end of the prompt.

\end{document}